\documentclass{article}

\usepackage[final]{neurips_2019}

\usepackage[utf8]{inputenc} 
\usepackage[T1]{fontenc}    
\usepackage{hyperref}       
\usepackage{url}            
\usepackage{booktabs}       
\usepackage{amsfonts}       
\usepackage{nicefrac}       
\usepackage{microtype}      

\pdfoutput=1

\RequirePackage[OT1]{fontenc}
\RequirePackage{amsthm,amsmath}
\usepackage{hyperref}
\usepackage{url}

\usepackage{dsfont} 
\usepackage{xspace}
\usepackage[usenames,dvipsnames]{xcolor} 

\usepackage{graphicx}

\usepackage{algorithm}
\usepackage{algorithmic}

\usepackage{multirow}
\usepackage{hhline}

\usepackage{booktabs}

\usepackage{capt-of}

\newtheorem{theorem}{Theorem}

\newtheorem{defn}[theorem]{Definition}

\def\eps{\ensuremath{\epsilon}\xspace}

\allowdisplaybreaks 

\def\sig{\ensuremath{\sigma}\xspace}

\def\eps{\ensuremath{\epsilon}\xspace}

\def\sig{\ensuremath{\sigma}\xspace}

\usepackage[export]{adjustbox}

\usepackage{pbox} 

\newcommand{\fr}[2]{ { \frac{#1}{#2} }}
\def\lt{\left}
\def\rt{\right}

\makeatletter
\newcommand{\vast}{\bBigg@{3}}
\newcommand{\Vast}{\bBigg@{4}}
\makeatother

\def\w{{{\boldsymbol w}}}

\def\la{\langle}
\def\ra{\rangle}

\usepackage{pifont}
\def\cD{\ensuremath{\mathcal{D}}\xspace}

\def\q{{{\boldsymbol q}}}

\def\w{{{\boldsymbol w}}}

\usepackage{enumitem}

\DeclareMathOperator{\Regret}{Regret}

\def\a{\ensuremath{\boldsymbol{a}}}

\def\R{\ensuremath{\mathbb{R}}} 
 
\def\E{\ensuremath{\mathbb{E}}}

\DeclareMathOperator{\EE}{\mathds{E}}

\DeclareMathOperator{\PP}{\mathds{P}}

\newcommand*\diff{\mathop{}\!\mathrm{d}}

\usepackage{prettyref}

\def\ddefloop#1{\ifx\ddefloop#1\else\ddef{#1}\expandafter\ddefloop\fi}
\def\ddef#1{\expandafter\def\csname c#1\endcsname{\ensuremath{\mathcal{#1}}}}
\ddefloop ABCDEFGHIJKLMNOPQRSTUVWXYZ\ddefloop
\def\ddef#1{\expandafter\def\csname #1#1\endcsname{\ensuremath{\mathbb{#1}}}}
\ddefloop ABCDEFGHIJKLMNOPQRSTUVWXYZ\ddefloop

\def\ddef#1{\expandafter\def\csname b#1\endcsname{\ensuremath{\boldsymbol{#1}}}}
\ddefloop abcdeghijklmnopqrstuvwxyz\ddefloop

\newcommand{\bxi}{{\boldsymbol{\xi}}}

\let\EE\undefined

\let\PP\undefined
\DeclareMathOperator{\EE}{\mathbb{E}}

\DeclareMathOperator{\PP}{\mathbb{P}}

\DeclareMathOperator{\erf}{\text{erf}}

\def\oned{{\normalfont{\text{1D}}}}

\def\hbg{\hat{\bg}}
\def\tM{\text{\normalfont{M}}}
\def\tD{\text{\normalfont{D}}}

\setcitestyle{authoryear,round,citesep={;},aysep={,},yysep={;}}
\ifdefined\nohyperref\else\ifdefined\hypersetup
\definecolor{mydarkblue}{rgb}{0,0.08,0.45}
\hypersetup{ %
  pdftitle={},
  pdfauthor={},
  pdfsubject={},
  pdfkeywords={},
  pdfborder=0 0 0,
  pdfpagemode=UseNone,
  colorlinks=true,
  linkcolor=mydarkblue,
  citecolor=mydarkblue,
  filecolor=mydarkblue,
  urlcolor=mydarkblue,
  pdfview=FitH}

\ifdefined\isaccepted \else
\hypersetup{pdfauthor={Anonymous Submission}}
\fi
\fi\fi

\usepackage[export]{adjustbox}

\usepackage{anyfontsize}

\usepackage{booktabs}       
\usepackage{amsfonts}       
\usepackage{nicefrac}       
\usepackage{microtype}      

\usepackage{mathtools}
\usepackage{bbm}
\usepackage{amsfonts}

\usepackage{MnSymbol} 

\mathtoolsset{showonlyrefs}

\usepackage{enumitem}

\title{Parameter-Free Locally Differentially Private Stochastic Subgradient Descent}

\author{
    Kwang-Sung Jun \\
    University of Arizona \\
    \texttt{kjun@cs.arizona.edu}\\
    \And
    Francesco Orabona\\
    Boston University \\
    \texttt{francesco@orabona.com}\\
}

\begin{document}

\maketitle

\defcitealias{cutkosky17online}{Cutkosky and Boahen (COLT 2017)}

\begin{abstract}%
  \vspace{-.3em}
  We consider the problem of minimizing a convex risk with stochastic subgradients guaranteeing $\epsilon$-locally differentially private ($\epsilon$-LDP).
  While it has been shown that stochastic optimization is possible with $\epsilon$-LDP via the standard SGD~\citep{SongCS13}, its convergence rate largely depends on the learning rate, which must be tuned via repeated runs.
  Further, tuning is detrimental to privacy loss since it significantly increases the number of gradient requests.
  In this work, we propose BANCO (Betting Algorithm for Noisy COins), the first $\epsilon$-LDP SGD algorithm that essentially matches the convergence rate of the tuned SGD without any learning rate parameter, reducing privacy loss and saving privacy budget.
\end{abstract}

\vspace{-1.5em}
\section{Introduction}
\vspace{-.5em}

In this paper, we consider the problem of minimizing the convex risk of a machine learning predictor, guaranteeing local differential privacy.
Instead of going through the empirical risk minimization route, we directly optimize the stochastic objective of the risk via stochastic subgradients appropriately sanitized to guarantee the local differential privacy. Our proposed algorithm is a variant of stochastic subgradient descent, with the important feature of \emph{not requiring the tuning of learning rates}. Yet, it guarantees the same convergence rate and local differential privacy of a stochastic subgradient descent procedure with the oracle tuning of its learning rate.

More in details, consider the risk $R(\bw):\R^d\rightarrow \R$ defined as $R(\bw):=\EE_{\bx\sim \rho_X}[\ell(\bw,\bx)]$, where $\ell(\bw,\bx)$ is convex in the first argument and $\bx$ represents sensitive data about an individual drawn i.i.d. from $\rho_X$. In machine learning terms, $R(\bw)$ represents the test loss that we are interested in directly minimizing, while $\ell(\bw,\bx)$ is the loss on a single sample $\bx$.
It is well-known that, by linearity of the expectation, we can get an unbiased estimate of a subgradient of $R(\bw)$ just using a subgradient w.r.t. the first argument of $\ell(\cdot,\bx)$ where $\bx$ is drawn i.i.d. from $\rho_X$. Hence, we can construct sanitized subgradients simply returning a noisy version $\cG(\bw) \in \partial \ell(\bw,\bx) + \bxi$ where the noise $\bxi$ guarantees the $\epsilon$-Local Differential Privacy ($\epsilon$-LDP), which prevents the private information from being identified.
We elaborate more on the problem definition and the assumptions in Section~\ref{sec:setting}.

One immediately obvious algorithm to try for $\eps$-LDP is the standard stochastic subgradient descent (SGD) algorithm together with the popular choice of density $p(\bxi) \propto \exp(-\tfrac{\eps}{2} \|\bxi\|)$~\citep{SongCS13,SongCS15,WuLKCJN17}.
Let us consider the SGD with a constant learning rate $\eta$ for simplicity.\footnote{The same argument holds for adaptive learning rates such as~\citet{DuchiHS10}.}
The convergence rate of SGD w.r.t. the optimum $\bw^* = \min_{\bw} R(\bw)$ after one pass over $T$ samples is $\EE[R(\bw_T)] - R(\bw^*) = O(  \fr{\|\bw^*\|^2}{\eta T} + \fr{d^2}{\eps^2}\eta)$.
Notice that the convergence rate largely depends on the learning rate, and setting $\eta = \fr{\eps/d}{\sqrt{T}}$ results in the rate of $O(\fr{d}{\eps}\|\bw^*\|^2/\sqrt{T})$.
While the optimal tuning $\|\bw^*\| \fr{\eps/d}{\sqrt{T}}$ would result in a much better rate of $O(\fr{d}{\eps}\|\bw^*\|/\sqrt{T})$, the value $\|\bw^*\|$ cannot be known beforehand.
Therefore, practitioners need to tune the learning rate by trying out various values over multiple runs.
However, this significantly increases the number of subgradient requests, which can be interpreted as the privacy cost.
This also implies that, when there is a fixed budget of privacy loss, one would have to use a smaller $\epsilon$ to allow multiple runs, which slows down the convergence.

Departing from the cumbersome and privacy-costing parameter tuning, we propose a new algorithm BANCO (Betting Algorithm for Noisy COins) that enjoys the convergence rate $\tilde O(\fr{d}{\eps}\|\bw^*\|/\sqrt{T})$ without having to tune any parameters, where $\tilde{O}$ hides logarithmic factors.
At the heart of the algorithm is a decomposition that isolates the work of adapting to the norm of the
optimum $\|\w^*\|$ and delegates it to a one-dimensional stochastic optimization algorithm.
We describe our algorithm and highlight the proof sketch in Section~\ref{sec:banach} and conclude with future directions in Section~\ref{sec:conclusion}.

\textbf{Related work.}~~
While there is a large body of work on differentially private (DP) empirical risk minimization, there is less research on studying the generalization properties of the obtained solution. Moreover, most of the work is based on the notion of differential privacy rather than on the stronger notion of local differential privacy (LDP). We remind the readers that the DP is useful when the data provider trusts the learning agents to be DP and allows them to access sensitive data (or data provider may possess the learners internally) whereas LDP is useful when the data provider does not trust the learning agents and thus have to release sanitized version of individual data.
Due to space constraints, we focus on discussing the most relevant works.
The notion of LDP was introduced in \citet{DuchiJW14} and \citet{SongCS13,SongCS15}. \citet{DuchiJW14} study the generalization of locally differentially private SGD, focusing on the optimal dependency on the number of dimensions, but assuming bounded domains to optimally tune the learning rates. \citet{SongCS13,SongCS15} use locally differentially private SGD to minimize the regularized ERM problems.
\citet{WuLKCJN17} study the convergence of SGD to minimize the ERM problem and guarantee differential privacy through the output perturbation method, and they also require the knowledge of the norm of the optimal solution to optimally tune the learning rates.
\citet{BassilyST14} studies the generalization properties of locally differentially private SGD, but it requires a bounded domain to set the learning rates or the presence of a regularizer, that in turn must be tuned.
Instead, our work gives a generalization guarantee for a locally differentially private SGD with one pass over the data and no learning rates to tune.

The problem of parameter-free $\eps$-LDP SGD can be seen as a noisy version of the parameter-free SGD problem~\citep{StreeterM12,OrabonaP16b,OrabonaT17,CutkoskyO18}.
However, the extension of existing (near-optimal) parameter-free SGDs to noisy gradients is not immediately clear since the noise makes the gradients unbounded while the parameter-free SGDs require bounded gradients.
Specifically, naively feeding in the unbounded gradients can even make the algorithm undefined, let alone the lack of convergence guarantees; see~\cite[Section 3]{JunO19} for a detailed argument.
Our solution in this paper goes back to the more general version of Online Convex Optimization (OCO)~\citep{Zinkevich03} and reconsider its formulation to allow unbounded noise, which is more general than the $\eps$-LDP setting.
In this paper, however, we focus on the LDP setting; we refer the interested readers to \cite{JunO19}.

\vspace{-.5em}
\section{Problem Definition and Preliminaries}
\label{sec:setting}
\vspace{-.5em}

In this section, we describe our notations and provide relevant background.

\vspace{-.8em}
\paragraph{Locally Differentially Private (LDP) SGD.}
An $\epsilon$-differentially private algorithm must guarantee that the log-likelihood ratio of the outputs of the algorithm under two databases differing in a single individual's data is smaller than $\epsilon$~\citep{DworkMNS06}. In the stricter definition of local differential privacy (LDP)~\citep{WassermanZ10,KasiviswanathanLNRS11,DuchiJW14,SongCS13,SongCS15} an untrusted algorithm is allowed to access a perturbed version of a sensitive data point only through a sanitization interface. In particular, the sanitization mechanism must guarantee that the log-likelihood ratio of the data of two individuals $x$ and $x'$ is smaller than $\epsilon$.
\begin{defn}[Local Differential Privacy]
  \label{def:local_diff_privacy}
  Let $D=(X_1, \dots, X_n)$ be a sensitive dataset where each $X_i \sim \rho_X$ corresponds to data about individual $i$.
  A randomized sanitization mechanism $M$ which outputs a disguised version $(U_1, \dots U_n)$ of $D$ is said to provide $\epsilon$-local differential privacy to individual $i$, if
  \[
  \sup_S \sup_{x,x' \in \cD} \frac{\PP[U_i\in S|X_i=x]}{\PP[U_i \in S| X_i = x']} \leq \exp(\epsilon),
  \]
  where the probability is w.r.t. the randomization in the sanitization mechanism.
\end{defn}
\vspace{-.2em}

Denote by $\partial \ell(\bw,\bx)$ the subdifferential set of $\ell$ w.r.t. its first argument evaluated at $(\bw,\bx)$.
The local differential setting can be specialized to SGD, as described in the introduction, by a mechanism that, upon receiving a subgradient in $\partial \ell(\bw,\bx)$ with $\bx \stackrel{i.i.d.}{\sim} \rho_X$, returns a noisy version $\cG(\bw) \in \partial \ell(\bw,\bx) + \bxi$, where the noise $\bxi$ guarantees the $\epsilon$-local differential privacy.

\vspace{-.8em}
\paragraph{Assumptions.} 
Throughout, we work with \textit{negative} subgradients for notational simplicity.
Let $\hbg_t$ be the negative subgradient received at time step $t$.
Let $\bg_t = \EE_t[\hbg_t]$ be the true negative subgradient and $\bxi_t = \hbg_t - \EE_t[\hbg_t]$ where we use the notation $\EE_t$ to denote $\EE [ \cdot \mid \bxi_{1}, \dots, \bxi_{t-1}]$.

We assume that the true subgradients are bounded by $G$: $\|\bg_t\|_2 \le G$.
Furthermore, the noise $\bxi_t$ is conditionally zero-mean and has conditional finite variance:
\begin{align}
\label{eq:bounded_variance}
\EE_t\lt[\|\bxi_t\|_2^2\rt] \le \sig^2, \forall t,
\end{align}
for some $\sig>0$.
We also assume a tail condition such that $\bxi_t$ is conditionally sub-exponential with parameters $(\sigma_\oned^2, b)$:
\begin{align}
\label{eq:sub_gamma}
\max_{\a: \|\a\|_2\le1} \ \E_t\left[\exp(\beta
\la \bxi_t, \ba \ra) \right] 
\leq \exp\left(\beta^2 \sigma_{\oned}^2/2\right), \  \forall |\beta| \leq 1/b~.
\end{align}
One can show that, when \eqref{eq:sub_gamma} is achieved with equality, we have $\sigma_{\oned}^2 \le \sigma^2$.
The intuition of the condition above is that the tail of the noise $\bxi_t$ behaves well in any direction.
This noise definition covers a wide range of distributions, including Gaussian and Laplace. 
If $d=1$, we have $\sig^2 = \sig^2_\oned$.
This is not true in general and the relationship depends on the noise distribution.
If $\bxi_t \sim \cN(0,s^2 \mathbf{I})$, then one can see that $\sig_\oned^2 = s^2$ and $\sig^2 = d s^2$.
As another example, the Laplace mechanism noise used in differentially-private learning satisfies the tail condition above; see \citet[Lemma 7]{JunO19}.

\vspace{-.5em}
\section{Parameter-free Stochastic Optimization with Noise}
\label{sec:banach}
\vspace{-.5em}

We now describe BANCO that achieves essentially the convergence rate of optimally-tuned SGD without repeated runs for tuning the learning rate, thus being truly one-pass.
BANCO decomposes the optimization process into two parts: one for optimizing the direction and the other for the magnitude of the solution vector.
For the direction, we  use projected stochastic gradient descent with adaptive learning rates~\citep{OrabonaP18}.
While, for the magnitude we use a parameter-free algorithm based on coin-betting~\citep{OrabonaP16b,JunO19}. 
We then construct the solution vector as the multiplication of the two solutions.
The full pseudocode can be found in Algorithm~\ref{alg:dpsgd}.
Note that $m_{t+1}$ has a closed form solution as follows: with shorthands $x = \sum_{s=1}^{t} \la \hat\bg_s, \bq_s\ra$ and $y= t(\sig^2/2 + G^2)$,
\begin{align*}
m_{t+1} = \fr{e^{-a(ay+x)}\lt( \sqrt\pi x  \exp\lt(\fr{(2ay+x)^2}{4y}\rt)       \lt( \erf(\fr{2ay+x}{2\sqrt y}) + \erf(\fr{2ay-x}{2\sqrt y}) \rt) + 2\sqrt y(1-e^{2ax}) \rt)}{8ay^{3/2}}~.
\end{align*}
For improving numerical precision for computing $m_{t+1}$, we refer to~\citep{koolenblog}.

\textfloatsep=.5em
\begin{algorithm}[t]
  \begin{algorithmic}[1]
    \STATE Set $\bw_1 = \bq_1 = \boldsymbol{0} \in \R^d$
    \FOR {$t=1$ \TO $T$}
    \STATE Receive a noisy negative subgradient $\hbg_t$ such that $\EE[\hbg_t] \in -\partial \ell(\bw_t, \bx)$ where $\bx \sim \rho_X$
    \STATE Update magnitude: $m_{t+1} = \frac{1}{2a} \int_{-a}^{a} \beta\exp\left(\beta \sum_{s=1}^{t} \la \hat\bg_s, \q_s\ra - \beta^2 t \left(\frac{\sig^2}{2} + G^2\right) \right)  \diff \beta$\\ where $a = \min\left(\fr{k_1}{G}, \fr{1}{b}\right)$, and $k_1=0.6838$
    \STATE Update direction: $\bq_{t+\frac{1}{2}} = \bq_t - \frac{\hbg_t}{\sqrt{\sum_{s=1}^t \|\hbg_s\|_2^2}}$
    \STATE Project direction onto $L2$ ball: $\bq_{t+1} = \bq_{t+\frac{1}{2}} \cdot \min\left(1, \left\|\bq_{t+\frac{1}{2}}\right\|_2^{-1}\right)$
    \STATE Update the weight vector: $\bw_{t+1} = m_{t+1} \bq_{t+1} \in \R^d$
    \ENDFOR
    \STATE{Return $\frac{1}{T}\sum_{t=1}^T \bw_t$}
  \end{algorithmic}  
  \caption{Betting Algorithm for Noisy COins (BANCO) for Differentially Private SGD}
  \label{alg:dpsgd}
\end{algorithm}

We now show that BANCO is a parameter-free $\eps$-LDP SGD algorithm.
Consider the Laplace sanitization mechanism that adds noise with probability density function $\rho_{\bxi}(\bz) \propto \exp(-\frac{\epsilon}{2} \|\bz\|_2)$, which makes the subgradients very similar to one another.
\citet{SongCS15} proved that this mechanism is $\epsilon$-locally differentially private and ensures $\EE\lt[\|\bxi_t\|_2^2\rt]\leq \frac{4(d^2+d)}{\epsilon^2}$, satisfying \eqref{eq:bounded_variance}.
Further, \citet[Lemma 7]{JunO19} show that this mechanism satisfies \eqref{eq:sub_gamma} with $\sig^2_\oned = 18d^2/\eps^2$ and $b=\eps/4$.

\vspace{-.3em}
\begin{theorem}
\label{thm:convergence_sgd}
Assume that $\ell(\bw,\bx)$ is convex in the first argument $\bw\in\RR^d$ and has its subgradients' L2-norm bounded by 1, where the subgradient is with respect to the first argument.
Let the noise $\hat\bg_t - \EE[\hat\bg_t]$ follow the density $\rho(\bz) \propto \exp(-\frac{\epsilon}{2}\|\bz\|_2)$.
Then, for any $\bw^\star \in \R^d$, after one pass over $T$ samples Algorithm~\ref{alg:dpsgd} guarantees 
\begin{align*}
\E \left[ R\left(\frac1T \sum_{t=1}^T \bw_t\right)\right] - R(\bw^\star)
\leq O\left( \frac{d\|\bw^\star\|_2}{\epsilon \sqrt{T}} \sqrt{\ln\left(1+\tfrac{d^2\|\bw^\star\|_2 T}{\epsilon^2}\right)} +\frac{1}{T}\right)~.
\end{align*}
\end{theorem}
This convergence rate matches the private SGD of \citet{WuLKCJN17} up to polylogarithmic terms, with the important difference that we do not need to assume the knowledge of the norm of the optimal solution $\bw^\star$ to tune the learning rates. 
Furthermore, the rate in Theorem~\ref{thm:convergence_sgd} is unimprovable up to logarithmic factors as shown in~\citet{JunO19}.
Finally, we remark that Algorithm~\ref{alg:dpsgd} is in fact an instance of~\citet{JunO19}, and more generic algorithms are available via a reduction therein, which can be of independent interest. Also, the Gaussian noise can be used as well, resulting a better dependency in the dimension of the space, but in the weaker $(\epsilon,\delta)$-LDP.

\vspace{-.3em}
\subsection{Proof Sketch}
\vspace{-.3em}

The proof is mainly concerned with regret, a standard quantity in online learning that measures the suboptimality with an arbitrary comparator $\bu$.
Theorem~\ref{thm:convergence_sgd} is a direct consequence of the expected regret bound we describe below using so-called the online-to-batch conversion~\citep{Cesa-BianchiCG04}.
Theorem~\ref{thm:banach-oco} below shows that the expected regret of Algorithm~\ref{alg:dpsgd} is decomposed into two expected regrets, each for the magnitude learner $m_t$ and the direction learner $\bq_t$.
The fact that we require the expected regret of the direction learner w.r.t. the \textit{unit norm comparator} frees us from tuning its learning rate, delegating the burden of adaptation to the magnitude learner. 
In turn, the magnitude algorithm is parameter-free, giving the best guarantee without having to tune any learning rates.
The proof is simple and immediate from \citet{CutkoskyO18}.
Hereafter, we consider linear functions that can be written as $\ell(\bw,\bx) = \la- \bg_t, \bw\ra$ for some $\bg_t$ without loss of generality (due to the convexity). 
\begin{theorem}
\label{thm:banach-oco}
  Suppose the direction learner obtains expected regret $\Regret^{\tD}_T(\bu) := \sum_{t=1}^T \la\hbg_t, \bu - \bq_t \ra$ for any competitor $\bu$ in the unit ball $S \subset \R^d$ and the magnitude learner obtains expected regret $\Regret^\tM_T(v) := \sum_{t=1}^T s_t\cdot(v - m_t)$ for any competitor $v\in\RR$ where $s_t = \la \hbg_t, \bq_t \ra$.
  Then, the iterates $m_t \bq_t$ guarantee
  \vspace{-.4em}
  \begin{align*}
    \EE \Regret_T(\bu) 
    :=  \EE \sum_{t=1}^T \la \bg_t, \bu - m_t\bq_t \ra 
    =\EE  \sum_{t=1}^T \la \hat\bg_t, \bu - m_t\bq_t \ra 
    \le \Regret^\tM_T( \|\bu\| ) + \|\bu\| \Regret^{\tD}_T(\tfrac{\bu}{\|\bu\|}),
  \end{align*}
  where we define $\bu/\|\bu\| = \boldsymbol{0}$ when $\bu = \boldsymbol{0}$.
\end{theorem}

For the direction learner, one can invoke any algorithm that guarantees the optimal regret with respect to competitors in the unit ball. Here, we have used projected online gradient descent with the scale-free learning rates in \citet{OrabonaP18}.
One can then immediately obtain the expected regret bound with noisy subgradients: 
\begin{align*}
  \EE \lt[ R_T^{\tD}\lt(\fr{\bu}{\|\bu\|_2}\rt) \rt]
  = O\lt(\EE \lt[ \sqrt{\sum_{t=1}^T \|\hbg_t \|_2^2} \rt] \rt) 
  \stackrel{(a)}{=} O\lt(\sqrt{\sum_{t=1}^T\lt( \EE \|\bg_t\|_2^2 + \sig^2 \rt)}\rt),
\end{align*}
where $(a)$ uses Jensen's inequality and the fact that $\EE[ \|\hbg_t\|^2] = \EE[\| \bg_t \|_2^2] + \sig^2$.

Finally, for the magnitude learner, we have instantiated the coin betting algorithm of \citet{JunO19} that has the following regret bound:
\begin{align*}
  \E [R_T^D(u) ]
  &= O\Bigg( |u| \max\lt\{(1+b) \ln \lt(|u| (1+b)\rt)   , \sqrt{(1 + \sig_\oned^2)T \ln\lt(|u|(1 + \sigma_\oned^2) T   +1\rt) } \rt\} + 1 \Bigg)~.
\end{align*}

\vspace{-.5em}
\section{Future Work}
\vspace{-.5em}
\label{sec:conclusion}

Our study opens up numerous research directions.
First, one immediate difference in our upper bound from the standard SGD algorithm with adaptive step size is that we do not have a data-dependent regret bound; we have $(G^2 +\sigma^2)T$ rather than $\E[\sum_{t=1}^T \|\hat{\bg}_{t}\|^2]$.
It would be interesting to investigate whether data-dependent bounds are possible.
Second, it would be desirable not to require the knowledge of the noise through $(\sig^2,b)$ so we can have a single algorithm that can adapt to various types of noise.
Finally, high probability regret bounds are a straightforward research direction.

\vspace{-.5em}
\subsection*{Acknowledgments}
\vspace{-.5em}
This material is based upon work supported by the National Science Foundation under grant no. 1740762 “Collaborative Research: TRIPODS Institute for Optimization and Learning.” We would like to thank Adam Smith for his valuable feedback on differentially-private SGDs.

\bibliographystyle{plainnat_nourl}
\bibliography{../../../../learning}

\end{document}